\documentclass[runningheads,a4paper]{llncs}
\pdfoutput=1
\usepackage{amssymb}

\usepackage{graphicx}
\usepackage[utf8]{inputenc}

\usepackage{url}
\urldef{\mailsa}\path|{M.Kursa,W.Rudnicki}@icm.edu.pl|
\urldef{\mailsb}\path|lukasz.komsta@am.lublin.pl|

\newcommand{\keywords}[1]{\par\addvspace\baselineskip
\noindent\keywordname\enspace\ignorespaces#1}

\begin{document}

\mainmatter  

\title{Random forest models of the retention constants in the thin layer chromatography}

\titlerunning{Random forest models of the retention constants in TLC}

%
%
\author{Miron B. Kursa\inst{1} \and Łukasz Komsta\inst{2} \and Witold R. Rudnicki\inst{1}}

\institute{Interdisciplinary Centre for Mathematical and Computational Modelling, University of Warsaw,\\
Pawinskiego 5a, Warsaw, Poland\\
\mailsa\\
\and
Department of Medicinal Chemistry, Skubiszewski Medical University,\\
Jaczewskiego 4, Lublin, Poland\\
\mailsb\\
}

%
%

\maketitle

\begin{abstract}
In the current study we examine an application of the machine learning methods to model the retention constants in the thin layer chromatography (TLC).
This problem can be described with hundreds or even thousands of descriptors relevant to various molecular properties, most of them redundant and not relevant for the retention constant prediction. 
Hence we employed feature selection to significantly reduce the number of attributes. 
Additionally we have tested application of the bagging procedure to the feature selection. 
The random forest regression models were built using selected variables. 
The resulting models have better correlation with the experimental data than the reference models obtained with linear regression. 
The cross-validation confirms robustness of the models.
\end{abstract}

\section{Introduction}
Thin layer chromatography is very important analytical procedure employed in chemistry and biochemistry for identification of compounds present in the samples collected in various circumstances.
It is widely used for example in the forensic studies to identify possible drugs and poisons collected in the field, for monitoring the reaction progress, preparative substance isolation and in many other situations. 
The procedure is performed as follows: first experimental sample is dissolved in a special solvent and put on one end of the thin layer of the adsorbent material spread on the chromatographic plate. 
Than, the solvent diffuses through the adsorbent by capillary action carrying dissolved molecules with it. 
Because of the numerous complex interactions between molecules, adsorbent and solvent, the effective speed of a molecule depends of its type; this way the compounds are finally spatially separated when the procedure ends.

One of the fundamental issues in chromatography is the relationship between the structure of a compound and its retention in a particular chromatographic system. 
The quantitative structure-retention relationship (QSRR) investigations are widely established and often used in prediction of a retention for new solutes, finding the most informative structure descriptors for retention explaining and checking their compliance with the molecular theory of the separation \cite{Kaliszan1992,Kaliszan1986}. 

The QSRR investigations regarding to the thin layer chromatography resulted in many equations able to predict the retention for only some groups of analogs. 
Most of them concern the retention in reversed-phase systems, where retention is strictly correlated with the lipophilicity of the solute \cite{Pyka2001,Wang1999,Heberger2007}. 
On the other hand the QSRR models useful for prediction of the retention on silica gel (where the retention mechanism is much more complicated) are relatively scarce.  
Recently several models on atomic contributions \cite{Komsta2007}, and substituent groups \cite{Komsta2008,Komsta2008a}  were proposed by Komsta, 
These models have reasonable predictive abilities for most studied systems. 

Nevertheless, there are several problems with general linear models that limit their predictive power in some difficult cases. 
The molecular systems are often described with hundreds, or even thousands of various descriptors, whereas the experimental data on retention for TLC systems is limited to hundreds of molecules at best. 
The standard statistical methods are developed for systems where the number of data points is at least one order of magnitude higher than number of descriptive variables. 
Furthermore, they don't work well for systems with large number of variables that may also be collinear.
The linear models are developed using an a priori selected set of variables that are only a small subset of all descriptors that can be used to describe the system. 
However, such selection, even performed by the expert, is arbitrary and not necessarily optimal.  

An alternative approach for modelling large experimental data sets with unknown relationships between descriptive variables and response variable has been developed by the machine learning community. 
Machine learning is in principle similar to statistical modelling, yet the generality is achieved by optimising the training process so that the model will work well on previously unseen data rather than constraining the number of model's degrees of freedom based on some assumptions.

The current study is devoted to exploring application of the random forest (RF) algorithm \cite{Breiman2001} for modelling retention constants in thin layer chromatography. 
To this end we selected two examples from the systems studied previously by Komsta \cite{Komsta2008}, and performed extensive tests of the methodology. 
We show that application of the random forest for modelling retention in TLC systems gives models that are robust, have better accuracy than their linear counterparts. 
Moreover, the feature selection algorithm that was applied to reduce noise, have also chosen variables that make sense from the chemists point of view.

\section{Materials and methods}
The data for retention coefficients in two TLC systems:  Chloroform:methanol (90:10) (TAD) and Chloroform:cyclohexane:acetic acid (4:4:2) (TAK) were used.  
The data sets comprised coefficients for 225 and 257 compounds in the case of the TAD  and TAK systems, respectively. 
Each compound was described with 1667 descriptors based on their chemical structure, atomic composition, presence of specific functional groups,  molecular graphs, etc., obtained with E-Dragon software \cite{Tetko2005}.
The retention constants used for training and validation of models were extracted from Clarke's almanach \cite{Moffat2004}.

The random forest is a machine learning algorithm proposed by Leo Breiman \cite{Breiman2001}; it is an ensemble method grouping CART models. 
Each tree in the ensemble is built on different bagging subsample (bag) of objects and each split of the tree is constructed as a best split obtained on a randomly selected subset of descriptive variables. 
All individual trees are weak classifiers, however they are also only weakly co-correlated within the ensemble. 
In a regression task the predictions of each CART are averaged to produce the final prediction. 
During the training of an RF model, an out of bag (OOB) error approximation is computed by measuring error rate for each object only using those trees for which given object was not present in their training sets (bags). 
Random forest also provides an estimation of variable importance computed from the estimated accuracy losses caused by nullifying information contained in a particular attribute.

\begin{figure}[t]
 	\centering
 	\includegraphics[width=1 \textwidth]{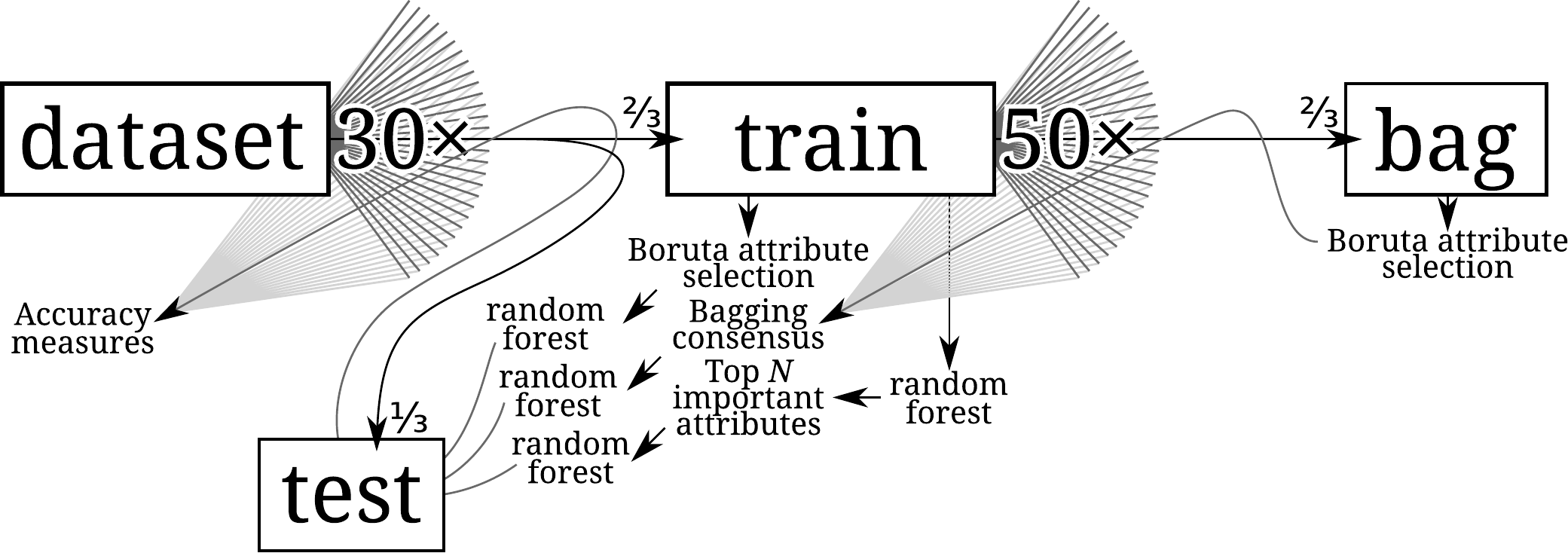}
	\caption{The scheme of the experimental procedure used in this study.}
 	\label{fig:ExpSummary}
 \end{figure}

The experimental procedure consisted thirty repetitions of two steps: feature selection followed by building a model using the reduced features set, see~Fig.~\ref{fig:ExpSummary}.
The following protocol was used for both data sets. 
The data set was randomly split in two parts, with proportion 2:1. 
The larger part was used for feature selection and model building, whereas the smaller part constituted a test set. 
Each step of the cross-validation loop started with a feature selection procedure. 
Once the relevant features were found the retention model was built using random forest algorithm \cite{Breiman2001}.
Finally the random forest model was used to predict the retention coefficients of the 1/3rd of compunds that were reserved as the test set and the OOB error of the model was compared with the predction error on the data that was previously put aside as a test. 

We used several feature selection algorithms. 
The first one is a simple minimal-optimal feature selection scheme proposed by Breiman in \cite{Breiman2001}. In this scheme one runs random forest twice, once with full set of features and with feature importance computations enabled and then repeats the run using the highest scoring features from the first run. 
We have run this procedure numerous times, with varying number of retained top scoring features. 
The procedure proved to be quite unstable in the cross-validation loop, returning changing sets of most important attributes in each iteration. 
In many cases even the top scoring attributes in one subset were not included in the most important list in most other subsets. 

The lack of stability of the feature selection procedure can influence the quality of models, therefore we used also two more elaborate schemes in hope of finding a small and stable feature set that would deliver consistent performance of random forest classifier across the cross-validation loop. 

In the first scheme we used a recently proposed algorithm for the all-relevant feature selection, Boruta,  \cite{Kursa2010a,Kursa2010c}.  
The algorithm is a heuristic procedure that utilises the importance measure provided by random forest to find all variables that are truly related with the decision. 
This is achieved by the use of contrast variables - the variables that are by design not correlated with the decision and have probability distributions similar to those of the original system. 
The algorithm deems important these attributes that in numerous random forest runs consistently have higher importance score than the best random contrast variables. 

The rationale for using the all-relevant feature selection is rather straightforward. 
In principle the results of the all-relevant feature selection should be independent of the sample, provided that sample is large enough to cover the variability in the population. 
In any case the set of important variables should be less variables than the ranking of variables within this set.  

The second, even more elaborate scheme employed in case if the stability of the all-relevant algorithm itself would not be sufficient. 
This scheme was based on the consensus variables, that is variables that were deemed important many times by the all-relevant feature selection algorithm executed on independent bagging samples of the original sample.
The rationale for this algorithm is following: assuming that the selection of objects to the sample does influence the result of the all-relevant feature selection algorithm, we then want to create a whole range of outcomes within each iteration of the cross-validation loop. 
In the ideal situation the variability within the sub-sample should be similar to that in the entire sample. 
On the other hand sub-sampling inevitably increases the noise and chance that the non-relevant attribute will be deemed relevant in some iterations. 
To deal with this noise only the variables that are deemed important in multiples trials are assumed to be really relevant. 

Thus a bagging procedure was performed at each step of the cross-validation loop. 
Fifty bagging samples were created from the training set, and the all-relevant feature selection procedure was applied to each of them. 
The consensus set $C_x$ comprised all attributes that were deemed important by the Boruta algorithm in at least $x \times N$ out of $N$ cases.   
The consensus feature sets $C_{0.1}$, $C_{0.2}$, \ldots, up to $C_{1.0}$ were created and used for building models. 



\begin{table}[htb]
\caption{Summary of feature selection results for the TopN algorithm. 
Each column holds number of attributes that were included in $N$ most important attributes at least K times, K is given in column head.  The last column holds the number of attributes that were included at most five times and can be considered noise. 
}
\begin{center}
\begin{tabular}{|r|r|r|r|r|r|r|r|r|}
  \hline
Common attributes & $=30$ & $\ge27$ & $\ge24$ & $\ge21$ & $\ge18$ & $\ge15$ & $\ge1$ & $\le5$ \\
  \hline
\multicolumn{9}{|c|}{TAD}\\
\hline
  TOP100 &  16 &  22 &  28 &  40 &  53 &  69  & 393 & 224 \\
  TOP90 &  16 &  22 &  26 &  37 &  47 &  62  & 368& 219  \\
  TOP80 &  14 &  19 &  26 &  29 &  43 &  54  & 331& 200 \\
  TOP70 &  12 &  18 &  25 &  28 &  35 &  48   & 302& 189 \\
  TOP60 &   9 &  18 &  21 &  26 &  32 &  38   & 250& 155 \\
  TOP50 &   7 &  17 &  20 &  22 &  28 &  34   & 202& 122 \\
  TOP40 &   3 &  14 &  17 &  21 &  23 &  29    & 160&  97  \\
  TOP30 &   1 &  10 &  15 &  19 &  20 &  23    & 125&  80 \\
  TOP20 &   1 &   5 &   8 &  13 &  15 &  18   &  74&  46 \\
  TOP10 &   1 &   1 &   1 &   5 &   7 &   8    &  42&  25 \\
   \hline
\multicolumn{9}{|c|}{TAK}\\
  \hline
  TOP100 &  11 &  27 &  42 &  53 &  59 &  70  & 443 & 295\\
  TOP90 &  11 &  25 &  38 &  47 &  56 &  66  & 397& 264 \\
  TOP80 &  11 &  23 &  33 &  44 &  54 &  59  & 333& 211 \\
  TOP70 &  10 &  20 &  28 &  42 &  46 &  55  & 283& 176\\
  TOP60 &   8 &  16 &  23 &  34 &  42 &  49  & 241 & 155\\
  TOP50 &   7 &  13 &  18 &  27 &  34 &  45  & 200 & 122\\
  TOP40 &   5 &  10 &  14 &  18 &  28 &  37  & 154 &  97\\
  TOP30 &   4 &   8 &  11 &  14 &  15 &  19  & 112&  63 \\
  TOP20 &   2 &   5 &   7 &   8 &  12 &  14  &  82&  48 \\
  TOP10 &   2 &   3 &   4 &   5 &   5 &   7  &  43&  28 \\
   \hline
\end{tabular}
\end{center}
\label{TAD-RF-IMP}
\end{table}

\section{Results}

\subsection{Feature Selection}

The simple feature selection procedure generates results that are highly non-replicable between different samples, as can be seen in Table~\ref{TAD-RF-IMP}, even though the classification results themselves are rather stable, as can be seen in Figure~\ref{fig:rsq}.
The number of attributes that are found in all or nearly all iterations is low. 
For example, only one attribute is consistently found across all 30 iterations in the TOP30 variant for TAD system, whereas 125 attributes were included at least once in the TOP30 set in any iteration. 
Moreover, only 5 attributes are found in at least 90\% of iterations and even at the consensus level as low as 50\%  only 23 commonly important attributes are found..

\begin{table}[htb]
\caption{Summary of feature selection results for the all-relevant algorithms. 
Columns holds number of attributes that were deemed important at least K times, K is given in column head.  
The last two columns hold the number of attributes that were included at most five times and can be considered noise and the average number of attributes deemed important, respectively.  
}
\begin{center}
\begin{tabular}{|r|r|r|r|r|r|r|r|r|r|}
  \hline
+Common attributes & $=30$ & $\ge27$ & $\ge24$ & $\ge21$ & $\ge18$ & $\ge15$ & $\ge1$ & $\le5$ & Average \\
  \hline
\multicolumn{10}{|c|}{TAD}\\
  \hline
  B1K & 2  & 10  & 14  & 18  & 23  & 23    & 246 & 188 & 41.77 \\
  B10K & 9  & 14  & 18  & 25  & 28  & 36    & 325 & 236 & 60.67 \\
  Bagging0.1 & 16  & 24  & 33  & 40  & 51  & 66   & 430 & 269  & 98.27 \\
  Bagging0.2 & 11  & 16  & 22  & 26  & 28  & 37    & 241 & 162 & 54.10 \\
  Bagging0.3 & 6  & 12  & 16  & 20  & 22  & 26    & 167 & 114 & 36.77 \\
  Bagging0.4 & 3  & 9  & 12  & 16  & 18  & 21    & 125 & 86 & 27.47 \\
  Bagging0.5 & 2  & 7  & 8  & 11  & 16  & 17   & 86 & 56  & 20.70 \\
  Bagging0.6 & 1  & 5  & 7  & 8  & 11  & 14 & 65 & 45    & 16.43 \\
  Bagging0.7 & 1  & 3  & 5  & 7  & 8  & 12 & 51 & 33    & 12.87 \\
  Bagging0.8 & 1  & 1  & 3  & 4  & 5  & 7  & 38 & 23    & 9.37 \\
  Bagging0.9 & 1  & 1  & 1  & 3  & 4  & 4 & 30 & 19   & 6.17 \\
  Bagging1 & 1  & 1  & 1  & 1  & 1  & 1 & 14  & 11   & 3.05 \\
  \hline
\multicolumn{10}{|c|}{TAK}\\
  \hline
  B1K & 6  & 12  & 17  & 26  & 32  & 37 & 171  & 103   & 45.07 \\
  B10K & 9  & 17  & 25  & 34  & 34  & 43 & 217 & 139    & 54.10 \\
  Bagging0.1 & 13  & 31  & 42  & 50  & 56  & 69 & 330  & 203   & 85.60 \\
  Bagging0.2 & 10  & 17  & 25  & 33  & 39  & 45  & 179 & 102    & 52.73 \\
  Bagging0.3 & 6  & 11  & 14  & 21  & 31  & 36 & 128 & 73    & 38.13 \\
  Bagging0.4 & 2  & 9  & 11  & 13  & 18  & 22  & 99 & 53    & 28.73 \\
  Bagging0.5 & 2  & 4  & 8  & 11  & 13  & 18 & 79  & 39   & 21.73 \\
  Bagging0.6 & 2  & 2  & 6  & 8  & 10  & 14  & 62 & 37    & 16.20 \\
  Bagging0.7 & 2  & 2  & 3  & 5  & 6  & 9  & 48  & 29   & 11.63 \\
  Bagging0.8 & 2  & 2  & 2  & 3  & 5  & 5 & 35 & 21    & 8.00  \\
  Bagging0.9 & 2  & 2  & 2  & 2  & 2  & 2  & 29 & 23    & 5.17 \\
  Bagging1 & 1  & 2  & 2  & 2  & 2  & 2 & 13  & 11   & 2.70 \\
   \hline
\end{tabular}
\end{center}
\label{All-relevant-selection}
\end{table}

The stability of feature selection does not influence highly the results of classification. 
Once the classifier has sufficiently large number of features to build model, the results are not improving despite increasing number of variables that are included in all iterations. 
Interestingly, it seems that the number of attributes that are relatively highly important in all sub-samples is limited. 
The number of attributes that appear in top $N$ attributes seems to saturate when $N$ approaches 100. 

Both algorithms using the all-relevant feature selection don't improve the stability of the feature selection process. 
The number of attributes that are deemed important in a given fraction of iterations is similar to the number of attributes that are returned by the $TopN$ algorithm returning similar number of top scoring attributes. 
The Boruta algorithm utilising random forests built from one thousand trees (Boruta 1K) is roughly equivalent to Top30 (for TAD system) and Top50 (for TAK system) algorithms, whereas Boruta 10K is roughly equivalent to Top60 algorithm for both systems, see Table~\ref{All-relevant-selection}.

The results obtained show that the feature selection process in both systems is strongly dependent on the sample. 
This arises due to  the high diversity of the data set. 
Consequently, if a training set contains too small few molecules belonging to similar class the classifier may rank lowly the variables best suited to describe these molecules, and hence they will not be included in the important set.  
This problem is fundamental and cannot be solved by using more sophisticated feature selection algorithm - simply the information present in the system is insufficient to deal with it. 

\subsection{Classification}

\begin{figure}[p]
 	\centering
 	\includegraphics[width=1 \textwidth]{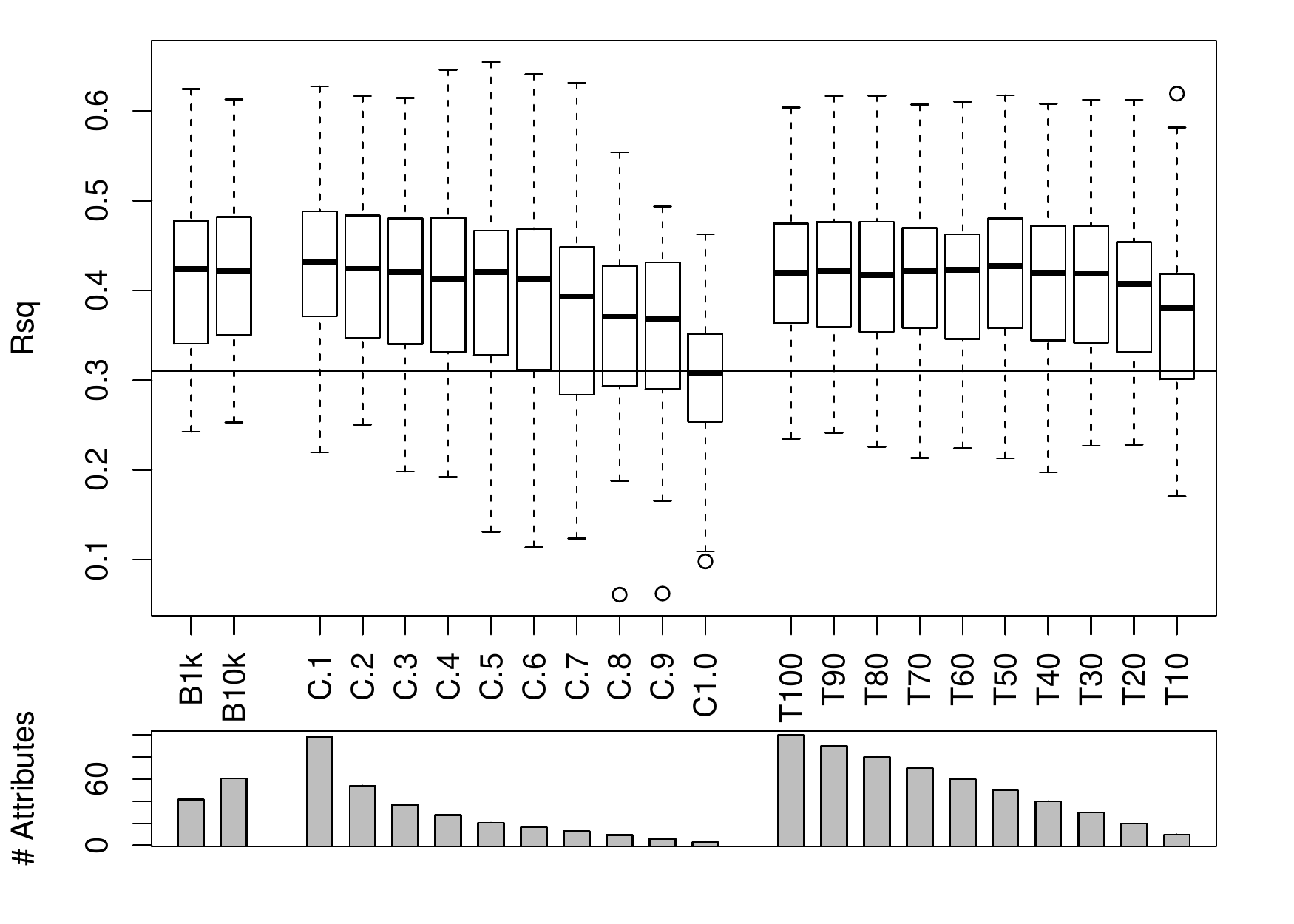}
	\includegraphics[width=1 \textwidth]{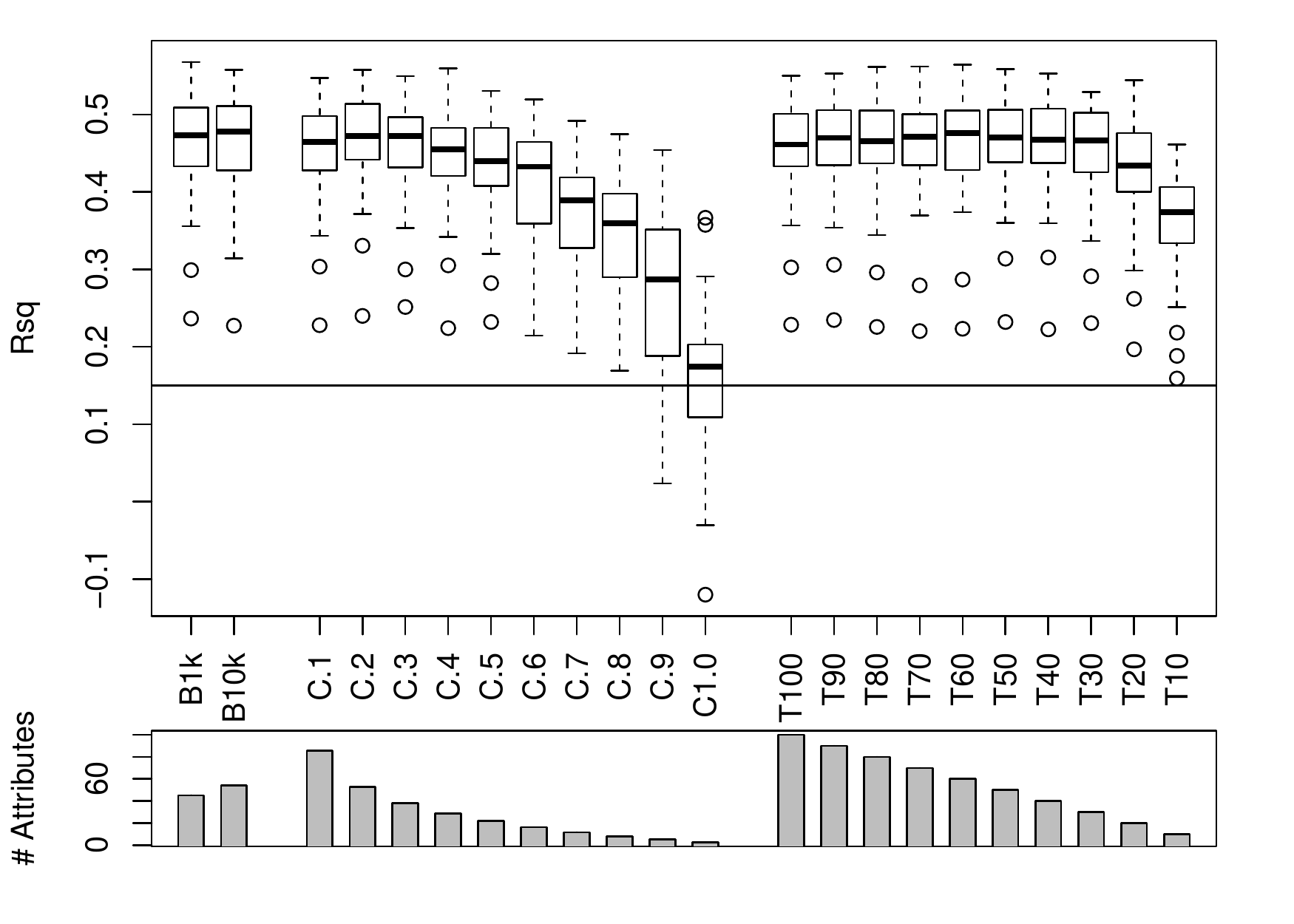}
	\caption{The cross-validated fraction of variance explained by the model of the TAD (top) and TAK (bottom) system, obtained by random forest classifier trained on the feature set returned by various feature selection algorithms. The horizontal line represents the baseline result reported from the linear model applied to the test set. The average number of attributes for each algorithm is shown in the lower box. }
 	\label{fig:rsq}
 \end{figure}

Despite irregular feature selection results the classification results are relatively good if the number of variables is sufficient, regardless of the feature selection algorithm used for constructing feature set, see Figure~\ref{fig:rsq}.

The results for TAD system are slightly better than the results reported earlier for the linear model. 
One may notice that even the model built on the feature set constructed with consensus algorithm that returns on average only three attributes are on par with linear model that was built using 9 variables. 
The models that are built with larger feature sets are significantly better. 
All models that were built using more than 10 attributes can explain on average at least 40\% of variance of the system. 
The best results, with nearly 43\% of variance explained by the model, were obtained using feature sets obtained with all-relevant Boruta algorithm as well as the $C_{0.1}$ consensus and $Top50$ algorithms, but the difference between models built using different feature sets are very small.    
These results are significantly better than the 31\% explained by linear model. 

The improvement over the linear model is more pronounced in the case of TAK system. 
Here the linear model built on 26 variables could explain only 15\% of the variance in the test system. 
The random forest models developed here can explain up to 48\% of variance for models using 60 variables. 
Nevertheless, the differences between models built on various feature sets were not high, provided that the feature set comprised 30 variables. 
Also in this case even the models built using few attributes (Top10, or high consensus all-relevant models) are better than the baseline  linear model. 
For example the models built using $C_{80\%}$ consensus set, which contains on average only 8 attributes, explain 36\% of variance in the data set, compared to 15\% for the linear model built on 36 variables. 
The best results, with over 47\% of variance explained by the model, were obtained using feature sets obtained with Boruta, $C_{0.2}$ and $Top60$ algorithms, but also here the differences between models are small.

\section{Discussion}
The experiments reported here have shown that the sample composition can significantly influence  the selection selection of relevant features in the case of modelling retention constant prediction in TLC.  
This problem cannot be overcome neither by using the all-relevant feature selection or even consensus methods. 
Most likely this issue originates in the difficulty of the problem rather than in the instability of a feature selection method itself.  
The attempt to model interactions of heterogeneous complex chemical entities certainly is not an easy task. 
Nevertheless, the random forest models improved significantly on the results published earlier with linear models.  
This results show utility of the machine learning methods for modelling the thin layer chromatography in particular and chemometry in general. 

In many cases the relations between descriptive and response variables are complex and non-linear, thus the  utility of linear models, even those including numerous variables, is limited. 
In comparison the random forest regression works very well, giving results that are significantly better and also more stable than those obtained with linear models. 
The question, whether application of other machine learning methods could improve on the current results will be explored in further studies, nevertheless, given that random forest is generally rather robust method no dramatic improvements are expected.

\bibliographystyle{plain}
\bibliography{text}

\begin{thebibliography}{10}

\bibitem{Breiman2001}
Leo Breiman.
\newblock {Random Forests}.
\newblock {\em Machine Learning}, 45:5--32, 2001.

\bibitem{Heberger2007}
K\'{a}roly H\'{e}berger.
\newblock {Quantitative structure -- (chromatographic) retention
  relationships.}
\newblock {\em Journal of chromatography. A}, 1158(1-2):273--305, 2007.

\bibitem{Kaliszan1986}
Roman Kaliszan.
\newblock {Quantitative relationships between molecular structure and
  chromatographic retention. Implications in physical, analytical, and
  medicinal chemistry}.
\newblock {\em Critical Reviews in Analytical Chemistry}, 16:323--383, 1986.

\bibitem{Kaliszan1992}
Roman Kaliszan.
\newblock {Quantitative structure retention relationships}.
\newblock {\em Analytical Chemistry}, 64:619--631, 1992.

\bibitem{Komsta2007}
Łukasz Komsta.
\newblock {Prediction of the retention in thin layer chromatography screening
  systems by atomic contributions.}
\newblock {\em Analytica Chimica Acta}, 593(2):224--37, 2007.

\bibitem{Komsta2008}
Łukasz Komsta.
\newblock {A functional-based approach to the retention in thin layer
  chromatographic screening systems.}
\newblock {\em Analytica Chimica Acta}, 629(1-2):66--72, November 2008.

\bibitem{Komsta2008a}
Łukasz Komsta.
\newblock {Quick prediction of the retention of solutes in 13 thin layer
  chromatographic screening systems on silica gel by classification and
  regression trees.}
\newblock {\em Journal of Separation Science}, 31(15):2899--909, 2008.

\bibitem{Kursa2010a}
Miron~B Kursa, Aleksander Jankowski, and Witold~R Rudnicki.
\newblock {Boruta -- A System for Feature Selection}.
\newblock {\em Fundamenta Informaticae}, 101(4):271--285, 2010.

\bibitem{Kursa2010c}
Miron~B Kursa and Witold~R Rudnicki.
\newblock {Feature Selection with the Boruta Package}.
\newblock {\em Journal Of Statistical Software}, 36(11), 2010.

\bibitem{Moffat2004}
A.C. Moffat.
\newblock {\em Clarke's Analysis of Drugs and Poisons}.
\newblock Pharmaceutical Press, 3 edition, 2004.

\bibitem{Pyka2001}
Alina Pyka.
\newblock {The application of topological indexes in TLC}.
\newblock {\em JPC. Journal of Planar Chromatography -- Modern TLC},
  14:152--159, 2001.

\bibitem{Tetko2005}
Igor~V Tetko, Johann Gasteiger, Roberto Todeschini, Andrea Mauri, David
  Livingstone, Peter Ertl, Vladimir~a Palyulin, Eugene~V Radchenko, Nikolay~S
  Zefirov, Alexander~S Makarenko, Vsevolod~Yu Tanchuk, and Volodymyr~V
  Prokopenko.
\newblock {Virtual computational chemistry laboratory--design and description.}
\newblock {\em Journal of computer-aided molecular design}, 19(6):453--63, June
  2005.

\bibitem{Wang1999}
QS~Wang and L~Zhang.
\newblock {Review of research on quantitative structure-retention relationships
  in thin-layer chromatography}.
\newblock {\em Journal of Liquid Chromatography \& Related Technologies},
  22(1):1--14, 1999.

\end{thebibliography}

\end{document}